\documentclass[preprint,12pt,sort&compress]{elsarticle}
\usepackage{bbding}
\usepackage{graphicx}
\usepackage{amsmath}
\allowdisplaybreaks[4]
\usepackage{array}
\usepackage{epsfig}
\usepackage{booktabs}
\usepackage{amssymb}
\usepackage{setspace}
\usepackage{bbding}
\usepackage{doi}
\usepackage{url}
\usepackage{algorithm}
\usepackage{algorithmic}
\usepackage{bbding}
\usepackage{multirow}
\usepackage{graphicx}
\graphicspath{ {./images/} }
\usepackage{caption}
\usepackage{subcaption}

\usepackage{enumitem}
\usepackage{tabularx}
\usepackage{amssymb}
\usepackage{amsmath}
\usepackage{afterpage}
\usepackage[figuresright]{rotating}

\usepackage{amssymb}

\begin{document}
\begin{frontmatter}



\title{GazeForensics: DeepFake Detection via \\ Gaze-guided Spatial Inconsistency Learning}


\author[1]{Qinlin He}
\ead{qinlin@stu.xidian.edu.cn}
\author[1]{Chunlei Peng\corref{mycorrespondingauthor}}
\cortext[mycorrespondingauthor]{Corresponding author}
\ead{clpeng@xidian.edu.cn}
\author[1]{Decheng Liu}
\ead{dchliu@xidian.edu.cn}
\author[2]{Nannan Wang}
\ead{nnwang@xidian.edu.cn}
\author[3]{Xinbo~Gao}
\ead{gaoxb@cqupt.edu.cn}
\address[1]{State Key Laboratory of Integrated Services Networks, School of Cyber Engineering, Xidian University, Xi'an 710071, Shaanxi, P. R. China}
\address[2]{State Key Laboratory of Integrated Services Networks, School of Telecommunications Engineering, Xidian University, Xi'an 710071, Shaanxi, P. R. China}
\address[3]{Chongqing Key Laboratory of Image Cognition, Chongqing University of Posts and Telecommunications, Chongqing 400065, P. R. China}

\begin{abstract}
DeepFake detection is pivotal in personal privacy and public safety. With the iterative advancement of DeepFake techniques, high-quality forged videos and images are becoming increasingly deceptive. Prior research has seen numerous attempts by scholars to incorporate biometric features into the field of DeepFake detection. However, traditional biometric-based approaches tend to segregate biometric features from general ones and freeze the biometric feature extractor. These approaches resulted in the exclusion of valuable general features, potentially leading to a performance decline and, consequently, a failure to fully exploit the potential of biometric information in assisting DeepFake detection. Moreover, insufficient attention has been dedicated to scrutinizing gaze authenticity within the realm of DeepFake detection in recent years. In this paper, we introduce \textit{GazeForensics}, an innovative DeepFake detection method that utilizes gaze representation obtained from a 3D gaze estimation model to regularize the corresponding representation within our DeepFake detection model, while concurrently integrating general features to further enhance the performance of our model. Experiment results reveal that our proposed \textit{GazeForensics} outperforms the current state-of-the-art methods.
\end{abstract}


\end{frontmatter}


\section{Introduction}

\begin{figure}[tp]
    \centerline{\includegraphics[width=0.55\textwidth]{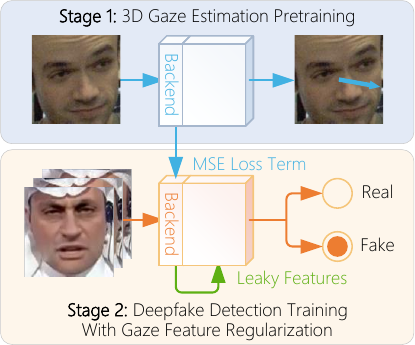}}
    \caption{By regulating the corresponding representation vector output by DeepFake detection backend with gaze feature vector while leaving a certain amount of features unconstrained, our DeepFake detection model can achieve a significant improvement in accuracy and robustness.}
    \label{fig:preview}
\end{figure}

In recent years, the evolution of facial manipulation technology has given rise to a formidable adversary. These techniques simplified the process of creating extraordinarily convincing face forgeries, amplifying the inherent challenges faced by traditional forgery detection methods. Deepfakes, along with feature editing, lip syncing, and so forth, with their capacity to seamlessly blend fabricated elements with genuine footage, have become potent tools for deception and manipulation, threatening both individuals and the whole society. Whether they are used to impersonate individuals, disseminate false narratives, or manipulate public perception, DeepFake technology poses a multifaceted threat that demands tailored solutions. In recognition of this urgent need, the development of precise and robust DeepFake detection algorithms is paramount.

Previously proposed DeepFake detection methods can be categorized into two groups based on whether they incorporate biometric features or not. Approaches that do not utilize biometric features
\cite{
    DFD_RNN_1,
    DFD_RNN_2,
    DCL,
    HCIL,
    F3_Net,
    Mesonet}
are designed to mine general cues that are helpful to DeepFake detection. These methods usually demonstrate superior robustness with respect to input data, as most of them don't require additional biometric feature extractors or other preprocessing steps. However, their explainability is slightly compromised compared to models that rely on biometric features since the exact evidence found by these models remains uncertain. In contrast, existing methods that employ biometric features
\cite{
    In_Ictu_Oculi,
    Exposing_Deepfake_by_Eye_Movements,
    Lips_Don_t_Lie,
    DeepRhythm,
    Where_Do_Deep_Fakes_Look}
typically utilize frozen pre-trained modules to extract desired biometric features as input for subsequent model components. These methods involve the selection of specific information relevant to certain biometric features, leading to the exclusion of general cues unrelated to the chosen biometric features.

To combine the aforementioned advantages of both biometric-based methods and non-biometric-based methods, we present our innovative gaze-based DeepFake detection method dubbed \textit{GazeForensics}, which utilizes gaze features to provide guidance and regularization for our DeepFake detection model. Figure \ref{fig:preview} depicts the basic concept of our proposed model. Observing that existing biometric-based DeepFake detection methods may be limited by incomplete data representations due to static backends and exclusion of general cues, we improved the training scheme and benefited from preserving general features besides biometric ones, balancing the biometric features and general features in a quantifiable way. Our approach is evaluated on the following datasets: FaceForensics++\cite{FFPP}, Celeb-DF\cite{Celeb_DF}, and WildDeepfake\cite{WDF}. The experimental results on both FaceForensics++ and WildDeepfake datasets demonstrate our superiority over current state-of-the-art (SOTA) approaches.

The contributions of our proposed approach can be summarized as follows:

\begin{enumerate}

\item We integrated DeepFake detection with 3D gaze estimation, filling this gap in the field of biometric-based DeepFake detection. This integration endows our model with the ability to discern forgery videos by distinguishing spatial inconsistencies within eye regions from a gaze perspective between given frames, improving the explainability, accuracy, and robustness of our model to some extent.

\item We proposed an innovative biometric feature integration strategy by introducing Mean Square Error (MSE) and leaky features fusion to regularize our DeepFake detection model. This feature fusion strategy not only provides a quantifiable method to balance general features and biometric ones, but also enhances the robustness and flexibility in handling challenging datasets.

\item Experimental results demonstrate that our proposed approach outperforms current SOTA models in both FaceForensics++ and WildDeepfake datasets and exhibits better explainability.

\end{enumerate}

The structure of this paper is as follows: Section 1 provides a brief introduction to the background and our proposed method. Section 2 presents frameworks and works associated with our proposed approach. In Section 3, we delve into the details of \textit{GazeForensics} framework. Section 4 showcases experimental results and their analysis. Finally, Section 5 concludes and summarizes our work.

\section{Related Work}

In this section, we will briefly overview the classical and recent methods for gaze estimation or DeepFake detection relevant to our work.

\subsection{Gaze Estimation}

Gaze estimation is an important research area that has applications in various fields, including psychology, medicine
\cite{
    Gaze_Cognitive_Styles_Study,
    Eye_tracking_research_in_eating_disorders,
    Gaze_ADHD}
, and human-computer interaction
\cite{
    Eye_Tracking_for_Everyone,
    Few_Shot_Personalization_for_Real_Time_Gaze_Estimation,
    Eye_Based_Human_Computer_Interaction,
    Gaze_Based_Applications}
. The ability to accurately estimate where a person is looking can provide valuable insights into their cognitive processes and behavior.
Unlike gaze following
\cite{
    Where_Are_They_Looking,
    Looking_Here_or_There,
    Following_Gaze_in_Video}
and 2D gaze estimation
\cite{
    Eye_Tracking_for_Everyone,
    Few_Shot_Personalization_for_Real_Time_Gaze_Estimation}
the task of 3D gaze estimation does not involve identifying the object or location that a person is looking at. Instead, it aims to derive the direction of a person's gaze from images or image sequences.
Zhang \textit{et al.} have proposed and improved methods for 3D gaze estimation based on monocular images
\cite{
    Appearance_based_gaze_estimation_in_the_wild,
    Mpiigaze}
, offering insights into combining local eye features with overall head pose features for subsequent research.
In a similar vein, Cheng \textit{et al.} observed the phenomenon of "two-eye asymmetry" and introduced the ARE-Net
\cite{Appearance_Based_Gaze_Estimation_Plus}
to fully exploit the role of binocular information in gaze estimation.
However, the need to detect eye positions and employ additional modules to generate latent vectors describing head pose hindered the progress of research and application on gaze estimation. Consequently, approaches utilizing the full-face region as direct input gained attention, leading to the emergence of numerous full-face gaze estimation methods.
Inspired by Krafka \textit{et al.} 's work
\cite{Eye_Tracking_for_Everyone}
, Zhang \textit{et al.} proposed a full-face gaze estimation method based on spatial weights mechanism
\cite{Full_Face_Appearance_Based_Gaze_Estimation}
. Kellnhofer \textit{et al.} made significant contributions by optimizing bidirectional LSTM capsules
\cite{Bi_LSTM}
using a pinball regression loss, facilitating 3D gaze estimation for continuous image sequences
\cite{Gaze360}
. Abdelrahman \textit{et al.} applied a linear combination of regression and classification losses separately for each angle, and their framework exhibited exceptional accuracy in single-image gaze estimation
\cite{L2CS_Net}
.

\subsection{DeepFake Detection}

The increasing threat posed by advancements in facial manipulation technology
\cite{
    Deepfakes,
    FaceSwap,
    Face2Face,
    FaceShifter,
    NeuralTexture,
    Expression_transfer_for_facial_reenactment}
, alongside the growing utilization of Generative Adversarial Networks (GANs)
\cite{
    A_Style_Based_GAN,
    Improving_StyleGAN,
    FSGAN,
    StarGAN,
    PGGAN}
in the realm of DeepFake, has garnered mounting attention from both the populace and researchers. Existing methods for detecting DeepFakes can broadly be categorized into two categories based on whether they rely on biometric features.

A wide spectrum of techniques has been employed in the realm of DeepFake detection methods that do not rely on biometric features. The work of Güera \textit{et al.} \cite{DFD_RNN_1} and the method proposed by Sabir \textit{et al.} \cite{DFD_RNN_2} both utilized a Recurrent Neural Network (RNN) to process features extracted by Convolutional Neural Networks (CNNs) in each frame, aiming to identify temporal inconsistencies in forged videos. To mine both spatial and temporal inconsistency within DeepFake videos, Gu \textit{et al.} proposed Spatial-Temporal Inconsistency Learning (STIL). Approaching DeepFake detection from a contrastive learning perspective also showed great potential. Sun \textit{et al.} employed two distinct modules to identify the associations and disparities between frames and, based on this, introduced the Dual Contrastive Learning (DCL) architecture \cite{DCL}. Gu \textit{et al.} took a different approach by delving into local and global contrast separately, presenting the Hierarchical Contrastive Inconsistency Learning (HCIL) framework \cite{HCIL}. Concerning content generated through GANs, prior research has analyzed image frequency domains using Deep Neural Networks (DNNs), leading to the development of numerous DeepFake detection methods based on image spectrum
\cite{
    gan_dct_anomalies,
    Detecting_artifacts_in_gan,
    Fourier_spectrum_discrepancies_in_gan_1,
    Fourier_spectrum_discrepancies_in_gan_2,
    F3_Net,
    HFC-MFFD}
. In addition to the aforementioned studies and methods derived from hand-crafted features in the early stages
\cite{
    Early_noise,
    Early_filter}
, several researchers have sought to enhance model performance by applying increasingly complex DNNs to the task of DeepFake detection
\cite{
    Resnet,
    Xception,
    Recasting_Resnet_2_Forgery_Detection,
    Capsule_network_fake_detect,
    Mesonet}
.

With regard to the DeepFake detection methods that incorporate  biometric features, they are explicitly designed for the identification and validation of certain biometric characteristics, such as blinking
\cite{
    DeepVision,
    In_Ictu_Oculi}
, heartbeat \cite{DeepRhythm} , mouth movements \cite{Lips_Don_t_Lie} , or eye movements
\cite{
    Binocular_Synchronization,
    Exposing_Deepfake_by_Eye_Movements,
    Where_Do_Deep_Fakes_Look}
. These methods employ well-designed frameworks to discern counterfeit content based on an in-depth understanding of these biometric features. Both the studies conducted by Li \textit{et al.} \cite{In_Ictu_Oculi} and Jung \textit{et al.} \cite{DeepVision} employed the feature of eye blinking as a clue for detecting DeepFake videos. In the former scenario, a CNN was employed to extract distinctive features from eye region images. Subsequently, Long Short-Term Memory (LSTM) cells \cite{LSTM} were used to model temporal dependencies within the feature sequences. In contrast, the latter study entailed a statistical analysis of the blink intervals within the samples, which was conducted in comparison to the data available in their reference database. DeepRhythm \cite{DeepRhythm} employed remote visual photoplethysmography to monitor the periodic changes of facial skin color caused by heartbeat, which is considered a distinctive feature that cannot be replicated by frame-by-frame generated DeepFake videos in their assumption. LipForensics \cite{Lips_Don_t_Lie} distinguished DeepFake contents by freezing and fine-tuning modules after pre-training them on the visual speech recognition task, focusing on mining high-level semantic irregularities in mouth movements. Wang \textit{et al.} concatenated blink sequences with gaze vector sequences \cite{Binocular_Synchronization}, emphasizing both binocular blinking and the consistency of binocular movements. Li \textit{et al.} defined four features related to eye movements and fed the extracted features into a support vector machine to identify DeepFake videos \cite{Exposing_Deepfake_by_Eye_Movements}. Demir \textit{et al.} conducted an in-depth analysis of differences between DeepFake and genuine videos regarding eye appearance, motion patterns, binocular consistency, and other features. They extracted these features and used them as inputs for subsequent binary classification DNN model \cite{Where_Do_Deep_Fakes_Look}, offering valuable insights for future researchers on distinguishing deepfake videos based on eye-related characteristics.

Many methods employing biometric features heavily rely on frozen pre-trained backends for feature extraction
\cite{
    Lips_Don_t_Lie,
    Binocular_Synchronization,
    Exposing_Deepfake_by_Eye_Movements}
, resulting in a notable loss of general information. Consequently, there is a demand for a DeepFake detection method that preserves a wide range of general information while capturing essential detailed biometric features.

\section{Proposed Approach}

In this section, we present \textit{GazeForensics}, a DeepFake detection method that integrates gaze-related features with general features to enhance the model's accuracy and robustness. Our approach primarily focuses on discerning spatial inconsistencies within eye regions between arbitrary different frames. This is achieved by incorporating regularization and expansion techniques into the representation vectors extracted by our model's backend, allowing for the quantification of gaze information's significance in the model's decision-making process.

\subsection{Overview}

\begin{figure}[tp]
    \centerline{\includegraphics[width=0.55 \textwidth]{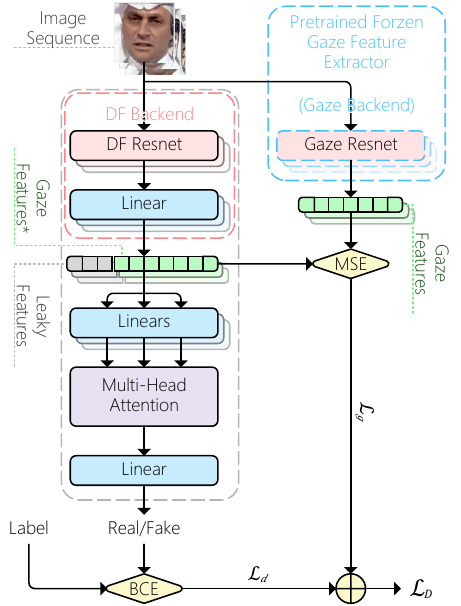}}
    \caption{The overview of \textit{GazeForensics}. Our framework utilizes a frozen pre-trained gaze feature extractor to extract gaze-related features, which are then employed to regularize the DeepFake detection backend through an MSE loss term. Simultaneously, a certain amount of features are allowed to bypass this regularization process, offering both general and supplementary features. In this context, the symbols $\oplus$, BCE, MSE, and Gaze Features* represent scalar addition, binary cross-entropy, mean squared error, and the gaze features estimated by the DeepFake detection backend, respectively.}
    \label{fig:modelStructure}
\end{figure}

In \textit{GazeForensics}, we aim to develop an accurate and robust DeepFake detection model. Our approach centers on the meticulous analysis of spatial inconsistency inherent in gaze-related features across disparate frames. Moreover, we harness supplementary general features to augment the model accuracy even further. We hypothesize that DeepFake videos, particularly those generated using frame-by-frame techniques, exhibit discernible discrepancies in preserving biometric attributes within the ocular regions \cite{Visual_Artifacts} such as iris color, reflection properties, and eye shape. These artifacts are deemed retrievable in gaze feature vector output by a 3D gaze estimation backend, offering valuable insights into the detection of potential manipulations.

In contrast to prior biometric-based DeepFake detection methods that are dedicated to preserving the model's reliance on specific biometric features by freezing pre-trained backend modules
\cite{
    Lips_Don_t_Lie,
    Binocular_Synchronization,
    Exposing_Deepfake_by_Eye_Movements}
, we encourage our DeepFake detection backend to learn actively to avoid reduced representational capacity stemming from potential covariate shift \cite{DatasetShift}. To be specific, we introduced an MSE loss term to regularize the DeepFake detection backend with the gaze-related features extracted by the frozen CNN backend from a pre-trained 3D gaze estimation model. This additional MSE loss term serves as a guide, enabling the Deepfake detection backend to incorporate both DeepFake-detection-related features and detailed gaze-related features within an integrated representation vector. This integration is achieved as the representation extracted by the MSE-constrained Deepfake detection backend is influenced by both the MSE gaze constraint and the Binary Cross-Entropy (BCE) DeepFake detection loss. Therefore, this representation, denoted as $\mathbf{r}_{d}$, is an integration of the gaze representation ($\mathbf{r}_{g}$) and a latent DeepFake-detection-related cue representation ($\mathbf{r}_{ld}$):
\[
\mathbf{r}_{d} = \mathbf{r}_{g} + \mathbf{r}_{ld}
\]

To further enhance the incorporation of supplementary general features in assisting Deepfake detection alongside gaze-related features, we expanded the dimension of the layer that is constrained by the MSE loss term. This extension vector is allowed to escape the MSE constraint, granting it the freedom to learn any feature. Our expectation is that these additional dimensions should serve as a representation of supplementary features that $\mathbf{r}_{ld}$ cannot accommodate.

As illustrated in Figure \ref{fig:modelStructure}, our proposed DeepFake detection framework comprises a CNN backend for DeepFake detection, a frozen CNN backend pre-trained on 3D gaze estimation, a multi-head attention module \cite{AttentionIsAllYouNeed}, and several linear layers. To be specific, we employed Resnet-18 \cite{Resnet} as both our DeepFake detection backend and gaze estimation backend. Rather than using RNNs, we chose the attention mechanism, which enables simultaneous comparison over all frames. This choice is grounded in the advantages that the attention mechanism offers when conducting spatial feature comparisons in DeepFake detection. It's worth noticing that the gaze estimation backend is only used in the training phase, which is designed to transfer its knowledge of extracting detailed gaze-related features to the DeepFake detection backend by the MSE constraint.

\subsection{Formulations}

Let the dataset used for DeepFake detection be denoted as $D_{d}$, and it is defined as $\{(x^{i}_{d}, y^{i}_{d})\}^{N_{d}}_{i=1}$. Here, $N_{d}$ represents the total number of video clips in the dataset, where each element $x^{i}_{d}$ refers to an individual video clip, and $y^{i}_{d}$ indicates whether the corresponding video clip is genuine or fake. Correspondingly, let the gaze estimation dataset, Gaze360 dataset \cite{Gaze360} to be specific, be denoted as $D_{g}$, and it is defined as $\{(x^{i}_{g}, y^{i}_{g})\}^{N_{g}}_{i=1}$. In this context, $N_{g}$ represents the overall size of the gaze estimation dataset, where each element $x^{i}_{g}$ corresponds to a full face image, and $y^{i}_{g}$ comprises the yaw and pitch angles representing the direction of gaze.

We utilize the loss function denoted as $CLS$, which was formulated by Ahmed \textit{et al.} \cite{L2CS_Net}, to optimize our 3D gaze estimation model. The expression for this loss function is defined as follows:
\[
\mathcal{L}_{G} = \frac{1}{N_{g}} \sum^{N_{g}}_{i=1} CLS(y^{i}_{g}, f_{g}(x^{i}_{g}, \theta_{gb}, \theta_{go}))
\]
where the 3D gaze estimation model $f_{g}$ is parameterized by two sets of parameters: the backend parameters $\theta_{gb}$ and other parameters $\theta_{go}$. This model can be decomposed into two constituent parts: $f_{gb}$, the backend modules of the gaze estimation model, and $f_{go}$, the subsequent portion of the network. Their relationship can be expressed as $f_{g} = f_{go} \circ f_{gb}$. Similarly, the DeepFake detection model $f_{d}$ can be represented as $f_{d} = f_{do} \circ f_{db}$, where $f_{db}$ and $f_{do}$ represent their corresponding counterparts.

To account for additional features that are not captured by the MSE gaze constraint during the training process of DeepFake detection, we define the representation extracted by the frozen pre-trained 3D gaze estimation backend as $\mathbf{r}^{i}_{g} \in \mathbb{R}^{\eta}$. This representation is equivalent to the expression $f_{gb}(x^{i}_{d}, \theta_{gb})$. Likewise, the representation obtained from the DeepFake detection backend is denoted as $\mathbf{r}^{i}_{d} \in \mathbb{R}^{\eta + \lambda}$ and is defined by $f_{db}(x^{i}_{d}, \theta_{db})$. To clarify, we consider $\lambda$ as a hyperparameter that governs the number of unconstrained dimensions, allowing the quantification of the additional features.

Since we aim to integrate DeepFake cues into the gaze feature vector, we establish the total loss function $\mathcal{L}_{D}$ for the DeepFake detection training stage. This loss function comprises two distinct components, denoted as $\mathcal{L}_{d}$ and $\mathcal{L}_{g}$ respectively:

\begin{enumerate}

\item \textbf{DeepFake Detection Component ($\mathcal{L}_{d}$)}: This component quantifies the disparity between the model's predictions and the actual labels. $\mathcal{L}_{d}$ encapsulates the primary objective of identifying DeepFake content. The representation for $\mathcal{L}_{d}$ is shown as follows:
\[
\mathcal{L}_{d} = \frac{1}{N_{d}} \sum^{N_{d}}_{i=1} BCE(y^{i}_{d}, f_{d}(x^{i}_{d}))
\]
where BCE is defined as:
\[
BCE(y, \hat{y}) = -\left[y \cdot \log(\hat{y}) + (1 - y) \cdot \log(1 - \hat{y})\right]
\]
\item \textbf{Gaze Constraint Component ($\mathcal{L}_{g}$)}: The gaze constraint component employs the MSE loss and serves as an auxiliary loss term. It leverages the representation of gaze features to guide and constrain the feature extraction process within DeepFake detection. This enhances the model's sensitivity to subtle ocular inconsistencies indicative of DeepFakes. The mathematical expression for $\mathcal{L}_{g}$ is as follows:
\[
\mathcal{L}_{g} = \frac{1}{N_{d}} \sum^{N_{d}}_{i=1} MSE(f_{gb}(x^{i}_{d}), \mathbf{r}|^{i}_{d_{1:\eta}})
\]
where MSE is defined as:
\[
MSE(y, \hat{y}) = \frac{1}{N}\sum_{0}^{N}(y-\hat{y})^{2}
\]

\end{enumerate}

In the formulation of the overall loss function $\mathcal{L}_{D}$, we introduce a scaling factor denoted as $\mu$. This factor serves to modulate the relative influence of the gaze-constraint loss term $\mathcal{L}_{g}$. By adjusting $\mu$, we can precisely control the model's sensitivity to gaze-related features and the impact of gaze constraint on the overall performance of DeepFake detection. The overall loss $\mathcal{L}_{D}$ is defined as:
\[
\mathcal{L}_{D} = \mathcal{L}_{d} + \mu \cdot \mathcal{L}_{g}
\]

\section{Experiments and Results}

In this section, we present a comprehensive evaluation of \textit{GazeForensics} on publicly available DeepFake detection datasets. We initiate our evaluation by scrutinizing the impact of two pivotal parameters on model performance, which are $\mu$ and $\lambda$. Analytical experiments are conducted with the WildDeepfake dataset, allowing us to gain insights into the impact on our model when changing the $\mu$ and $\lambda$. Subsequently, we proceed to assess the effectiveness of our proposed approach on several datasets, including FaceForensics++, CelebDF, and WildDeepfake. We also compared our method against a spectrum of classical and recent DeepFake detection methods to provide a comprehensive perspective on its performance in diverse contexts. Finally, to substantiate the efficacy of our proposed approach, we conduct a series of ablation experiments, systematically investigating the contributions of various components.

\subsection{Datasets}
For our experimental evaluations, we employed three publicly accessible DeepFake detection datasets, each with distinct characteristics:

\begin{enumerate}

\item \textbf{FaceForensics++} \cite{FFPP}: The FaceForensics++ (FF++) dataset serves as an established benchmark for evaluating various DeepFake detection methods. It encompasses a wide range of forgery techniques, including Deepfakes\cite{Deepfakes} (DF), Face2Face\cite{Face2Face} (F2F), FaceSwap\cite{FaceSwap} (FS), NeuralTextures\cite{NeuralTexture} (NT), and so on. However, in our experiment, we followed the prevalent practice used by researchers working with the FF++ dataset by exclusively using DeepFake videos generated from the aforementioned four methods, alongside authentic videos. The dataset itself comprises 1000 real videos and 1000 manipulated videos for each manipulation type.

\item \textbf{Celeb-DF} \cite{Celeb_DF}: The Celeb-DF dataset offers a collection of high-quality videos. It selects celebrity interview videos from YouTube and employs an improved DeepFake synthesis algorithm for face manipulation. This dataset includes 590 authentic videos alongside 5639 DeepFake videos. Additionally, there are 300 authentic videos collected from YouTube that don't have corresponding DeepFake counterparts.

\item \textbf{WildDeepfake} \cite{WDF}: The WildDeepfake (Wild-DF) dataset presents a significant challenge with its diverse, internet-sourced content. It includes video samples of various qualities, undisclosed forgery techniques, and some videos with partial manipulation, making the ground truth labels less conclusive. This dataset consists of 3805 authentic video sequences and 3509 manipulated video sequences. Notably, the sequences in this dataset might represent different segments clipped from the same source video.

\end{enumerate}

For 3D gaze estimation pre-training, we utilized the Gaze360\cite{Gaze360} dataset. This dataset is distinguished by a broad spectrum of gaze and head poses, a variety of indoor and outdoor capture environments, and a diverse range of characters. These attributes make it well-suited for our pre-training purposes. The angular annotations of yaw and pitch indicate which direction the subject is looking in 3D space with respect to the camera origin. These data are measured using the standard AprilTag-based procedure \cite{AprilTag}. The dataset is intended to be used for developing and evaluating gaze-tracking models that can estimate 3D gaze direction accurately in unconstrained images.

\subsection{Experimental Settings}

We implemented \textit{GazeForensics} using the PyTorch framework, conducting all experiments on a single RTX 4080 GPU.

\textbf{Pre-training}: To bolster the robustness of the gaze estimation backend, we introduced a preliminary step involving 3D gaze estimation before commencing the DeepFake detection training. For our gaze estimation model, we adopted L2CS-Net\cite{L2CS_Net}. During the training of this model, we applied additional data augmentation techniques that were not originally included in the reference paper. These augmentations encompassed resolution randomization, color jitter, as well as random cropping and rotation. Resolution randomization deliberately reduced the training data resolution to random values. These augmentations were vital not only for addressing video quality variations in some DeepFake detection datasets but also for mitigating potential covariate shifts when transitioning to the DeepFake detection dataset. Following the refinement of our gaze estimation model to handle diverse image contents effectively, we utilized its backend in the subsequent DeepFake detection training phase.

\textbf{Preprocess}: We employed RetinaFace\cite{RetinaFace} for facial alignment in the datasets that require image cropping.
This cropping operation aims to achieve a uniform aspect ratio for all images and eliminate extraneous background information.
Subsequently, we divided each original video into sequences of 14 frames, resizing them to a standardized resolution of 224*224 pixels before inputting them into our models.
Notably, in the DeepFake detection training stage, the only applied data argumentation is a random horizontal flip since we mainly focus on model design.
Additionally, our dataset partitioning followed the recommended guidelines and specifications provided in the dataset papers and associated documentation.

\textbf{Training}: The model underwent 50 epochs of training with the AdamW optimizer employing a weight decay coefficient of 0.01. A maximum learning rate of 7e-5 was selected, and the OneCycleLR scheduler was used to introduce warm-up phases and control the learning rate during training. Throughout the training, preprocessed frames were jointly processed by both the \textit{GazeForensics} model and the frozen backend of L2CS-Net. To shift the model's focus from gaze-related features to DeepFake detection in later stages, the hyperparameter $\mu$ was gradually reduced. After each epoch, $\mu$ was multiplied by a factor of 0.912, eventually reaching 0.01 times its initial value after 50 epochs. This dynamic reduction of $\mu$ ensured that the DeepFake detection backend prioritized the DeepFake detection loss $\mathcal{L}_{d}$ in the overall loss function $\mathcal{L}_{D}$. This parameter schedule facilitated a balanced transition from extracting gaze-related representations to enhancing DeepFake detection capabilities within the DeepFake detection backend. Notably, this scheduling did not lead to rebounds in the value of $\mathcal{L}_{g}$ during our experiments, signifying a refined balance and adaptation as previously mentioned.

\begin{figure}[tp]
    \centerline{\includegraphics[width=0.55 \textwidth]{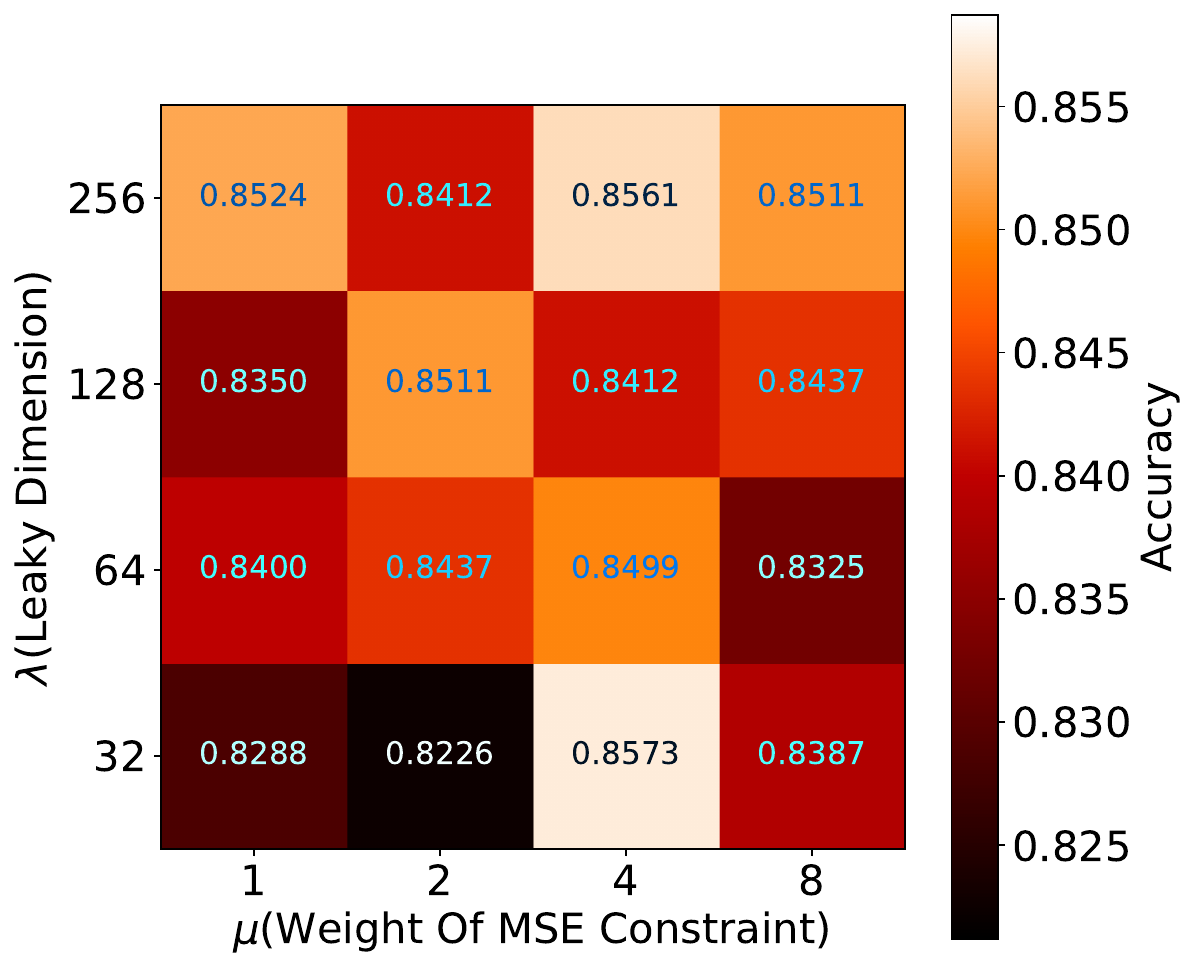}}
    \caption{Grid search result on Wild-DF dataset, accuracy on the test set are reported.}
    \label{fig:gridSearch}
\end{figure}

\textbf{Parameters}: We selected crucial hyperparameters, $\mu$ and $\lambda$, with grid search. To expedite the parameter search process within time and computational constraints, we predefined a range of potential values for each parameter. For $\mu$, we considered a range of four potential values: 1.0, 2.0, 4.0, and 8.0. Similarly, for $\lambda$, we confined our search to four predefined values: 32, 64, 128, and 256. To ensure result precision and highlight the varying impacts of different parameter choices, we conducted grid search experiments on the challenging WildDeepfake dataset. The experimental configuration aligns with the settings mentioned earlier in this subsection. Figure \ref{fig:gridSearch} illustrates the outcomes, with different $\mu$ and $\lambda$ combinations showing significant variability in accuracy. The observed accuracy ranged from a minimum of 0.8226 to a maximum of 0.8573. Notably, the parameter pair $\mu = 4.0$ and $\lambda = 32$ outperformed most other combinations in terms of test set accuracy. Therefore, these values were adopted as the default configuration for the subsequent experiments.

\subsection{Baseline Comparison}

In this subsection, we will compare our proposed method with recent or classical DeepFake detection approaches to demonstrate the superiority of our approach, including Xception \cite{Xception}, LSTM \cite{LSTM}, I3D \cite{I3D}, C3D \cite{C3D}, TEI \cite{TEI}, ADDNet-3D \cite{WDF}, F3-Net \cite{F3_Net}, S-MIL \cite{S-MIL}, STIL \cite{STIL}, DeepRhythm \cite{DeepRhythm}, ISTVT \cite{ISTVT}.

\begin{table}
    \centering
    \begin{tabular}{ccccccc}
    \toprule
    
        \multirow{2}*{Method}&  &  \multicolumn{5}{c}{FaceForensics++ c23 (HQ)}\\
        
        \cmidrule{3-7}
        
        &  &  DF&  F2F&  FS&  NT&  Avg.\\
        
    \midrule
    
        Xception \cite{Xception}&  &  0.9893&  0.9893&  0.9964&  0.9500&  0.9813\\

        C3D \cite{C3D}&  &  0.9286&  0.8857&  0.9179&  0.8964&  0.9072\\
        
        I3D \cite{I3D}&  &  0.9286&  0.9286&  0.9643&  0.9036&  0.9313\\
        
        LSTM \cite{LSTM}&  &  \textbf{0.9964}&  0.9929&  0.9821&  0.9393&  0.9777\\

        TEI \cite{TEI}&  &  0.9786&  0.9714&  0.9750&  0.9429&  0.9670\\
        
        DeepRhythm \cite{DeepRhythm}&  &  0.9870&  0.9890&  0.9780&  -&  -\\
        
        S-MIL \cite{S-MIL}&  &  0.9857&  0.9929&  0.9929&  0.9571&  0.9822\\
        
        S-MIL-T \cite{S-MIL}&  &  \textbf{0.9964}&  0.9964&  \textbf{1.0}&  0.9429&  0.9839\\
        
        STIL \cite{STIL}&  &  \textbf{0.9964}&  0.9928&  \textbf{1.0}&  0.9536&  0.9857\\
        
        ISTVT \cite{ISTVT}&  &  \textbf{0.9964}&  0.9964&  \textbf{1.0}&  0.9676&  0.9901\\
        
    \midrule
    
        GazeForensics(Ours)&  &  0.9893&  \textbf{1.0}&  \textbf{1.0}&  \textbf{0.9964}&  \textbf{0.9964}\\
        
    \bottomrule
    \end{tabular}
    \vspace{4pt}
    \caption{Comparison on high-quality FaceForensics++ dataset. Accuracy is reported.}
    \label{tab:Comp_FF++_HQ}
\end{table}

\begin{table}
    \centering
    \begin{tabular}{ccccccc}
    \toprule
    
        \multirow{2}*{Method}&  &  \multicolumn{5}{c}{FaceForensics++ c40 (LQ)}\\
        
        \cmidrule{3-7}
        
        &  &  DF&  F2F&  FS&  NT&  Avg.\\
        
    \midrule
    
        Xception \cite{Xception}&  &  0.9678&  0.9107&  0.9464&  0.8714&  0.9241 \\

        C3D \cite{C3D}&  &  0.8929&  0.8286&  0.8786&  0.8714&  0.8679 \\
        
        I3D \cite{I3D}&  &  0.9107&  0.8643&  0.9143&  0.7857&  0.8688 \\
        
        LSTM \cite{LSTM}&  &  0.9643&  0.8821&  0.9429&  0.8821&  0.9179 \\

        TEI \cite{TEI}&  &  0.9500&  0.9107&  0.9464&  0.9036&  0.9277 \\
        
        F3-Net \cite{F3_Net}&  &  0.9862&  0.9584&  0.9723&  0.8601&  0.9443 \\
                
        S-MIL \cite{S-MIL}&  &  0.9679&  0.9143&  0.9464&  0.8857&  0.9286 \\
        
        S-MIL-T \cite{S-MIL}&  &  0.9714&  0.9107&  0.9607&  0.8679&  0.9277 \\
        
        STIL \cite{STIL}&  &  0.9821&  0.9214&  0.9714&  0.9178&  0.9482 \\
        
        ISTVT \cite{ISTVT}&  &  \textbf{0.9893}&  0.9607&  0.9750&  0.9214&  0.9616 \\
        
    \midrule
    
        GazeForensics(Ours)&  &  0.9786&  \textbf{0.9964}&  \textbf{0.9964}&  \textbf{1.0}&  \textbf{0.9929} \\
        
    \bottomrule
    \end{tabular}
    \vspace{4pt}
    \caption{Comparison on low-quality FaceForensics++ dataset. Accuracy is reported.}
    \label{tab:Comp_FF++_LQ}
\end{table}

\textbf{Comparasion on FF++ dataset.}
We conducted comprehensive experiments on four sub-datasets of the FF++ dataset, encompassing two different video quality levels, Low Quality (LQ, c23) and High Quality (HQ, c40). Table \ref{tab:Comp_FF++_HQ} and Table \ref{tab:Comp_FF++_LQ} provide an overview of the accuracy of both previous methods and our proposed \textit{GazeForensics} on these datasets. In general, \textit{GazeForensics} achieves SOTA results for three types of forgery methods: Face2Face \cite{Face2Face}, FaceSwap \cite{FaceSwap}, and NeuralTexture \cite{NeuralTexture} in both LQ and HQ settings. Notably, our method excels in detecting manipulations created by NeuralTexture, demonstrating a remarkable 7.86\% increase in accuracy on the low-quality NerualTexture subset compared to previous SOTA methods. From the perspective of video quality, unlike previous methods that experienced significant accuracy degradation when transitioning from HQ to LQ, our approach maintains consistent accuracy levels across the F2F, FS, and NT manipulation types. This consistency showcases the superior adaptability of our method and underscores the robustness of our proposed framework concerning video quality. This may be attributed to the incorporation of random-resolution data augmentation during the pre-training of the gaze estimation backend and the integration of gaze-related information in the feature extraction. However, when compared to previous methods that consider Deepfakes as the easiest manipulation type to detect, \textit{GazeForensics} exhibits a slightly inferior performance, with a decrease of 1.07\% in detecting these manipulations.

\textbf{Comparasion on other datasets.}
We have also conducted experiments on the Celeb-DF and WildDeepfake datasets to demonstrate the superiority of \textit{GazeForensics}. As shown in Table \ref{tab:Comp_CDF_WDF}, our proposed method outperforms previous SOTA methods on the Wild-DF dataset. Given that the Wild-DF dataset is notably challenging and complex, the superior accuracy of our approach demonstrates the effectiveness of our unique combination of gaze constraint and general feature fusion in enhancing the model's generalizability and performance in complex scenarios. For the Celeb-DF dataset, which contains negative samples generated using advanced DeepFake forgery techniques, \textit{GazeForensics} leveraged the dataset's high visual quality and sophisticated forgery methods. This allowed for a detailed examination of spatial inconsistencies within the eye regions across image sequences. As a result, \textbf{GazeForensics} significantly narrowed the performance gap with previous methods to just 0.39\% on the Celeb-DF dataset. This gap is notably smaller, especially when compared to the DeepFakes presented in the FF++ dataset.

\begin{table}
    \centering
    \begin{tabular}{ccc}
    \toprule
    
         Method&  Celeb-DF& Wild-DF\\
         
    \midrule
    
        Xception \cite{Xception}&  0.9944&  0.8325 \\
        
        I3D \cite{I3D}&  0.9923&  0.6269 \\
        
        ADDNet-3D \cite{WDF}&  -&  0.6550 \\
        
        LSTM \cite{LSTM}&  0.9573&  -\\
        
        F3-Net \cite{F3_Net}&  0.9595&  0.8066 \\
        
        S-MIL \cite{S-MIL}&  0.9923&  - \\
        
        S-MIL-T \cite{S-MIL}&  0.9884&  - \\
        
        STIL \cite{STIL}&  0.9961&  0.8462 \\
        
        ISTVT \cite{ISTVT}& \textbf{0.9981}& - \\

    \midrule

        GazeForensics(Ours)&  0.9942&  \textbf{0.8573} \\
        
    \bottomrule
    \end{tabular}
    \vspace{4pt}
    \caption{Comparison on Celeb-DF and WildDeepfake datasets. Accuracy is reported.}
    \label{tab:Comp_CDF_WDF}
\end{table}

\subsection{Ablation Study}

In this subsection, we conduct an investigation into the effectiveness of the MSE gaze constraint and the utilization of leaky features through a series of meticulously designed experiments.

\begin{figure}
    \centerline{\includegraphics[width=0.5 \textwidth]{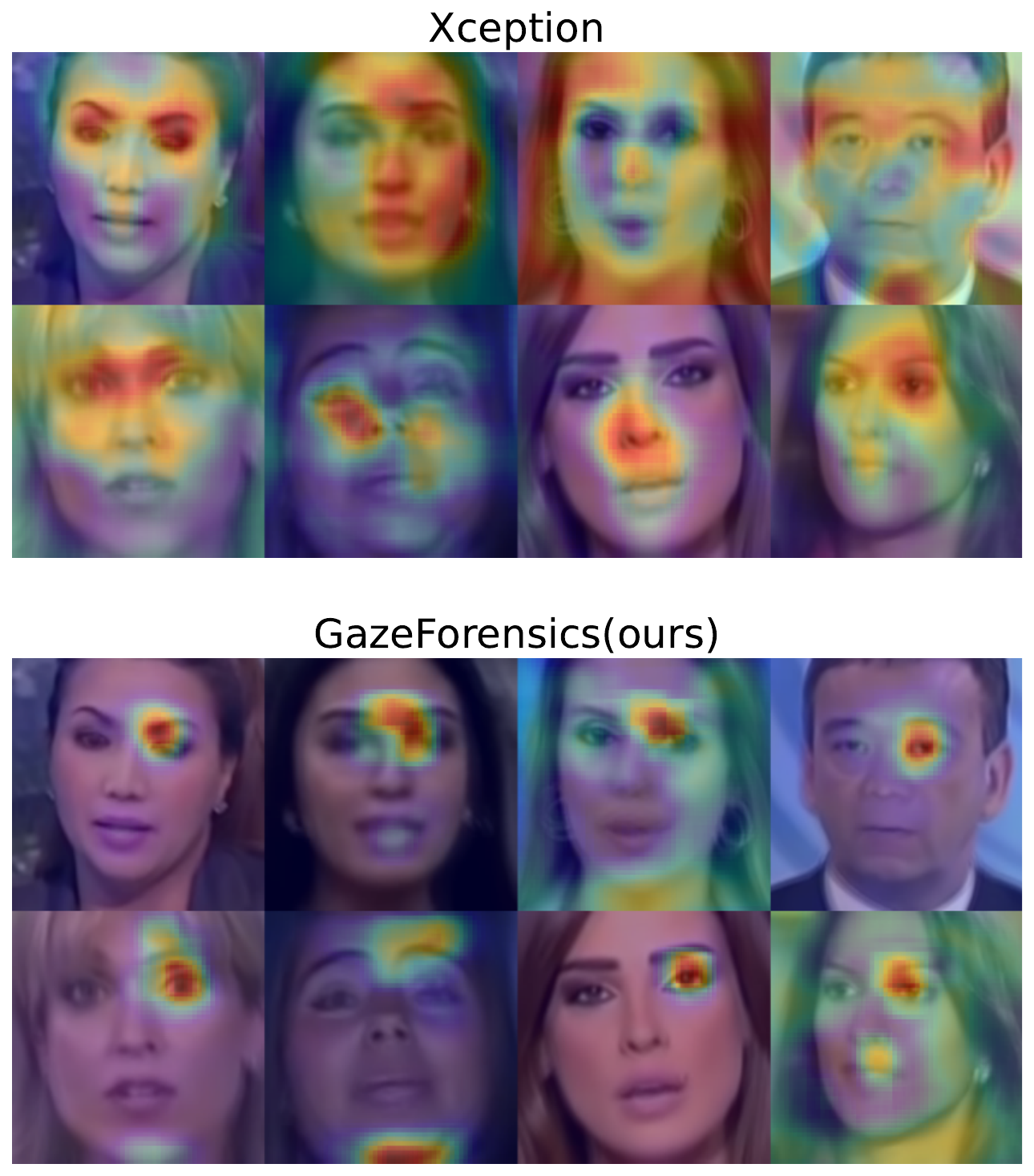}}
    \caption{Occlusion sensitivity samples measured with Xception and \textit{GazeForensics} on FF++ dataset}
    \label{fig:occSen}
\end{figure}

\textbf{Occlusion Sensitivity Visualization}: In our endeavor to assess the impact of incorporating gaze constraint on the decision-making process of the model, we conducted an experiment to visualize and compare the occlusion sensitivity of both \textit{GazeForensics} and the Xception \cite{Xception}. A randomly selected subset of test examples from the FF++ low-quality dataset underwent occlusion sensitivity measurement. Specifically, we introduced a grey patch to evaluate the models' responses under various occluded positions. The occlusion sensitivity visualization results are presented in Figure \ref{fig:occSen}. Upon visual inspection, it becomes evident that the occlusion sensitivity results observed in the Xception model exhibit no discernible pattern. In contrast, \textit{GazeForensics} places a strong emphasis on scrutinizing the eye regions to identify manipulation cues. Furthermore, it also demonstrates capability in utilizing more generalized cues in some samples, thereby expanding its versatility. These visualizations offer compelling evidence that gaze constraint significantly reshapes the foundational decision-making process in the DeepFake detection model, setting it apart from traditional methods. Consequently, this augmentation enhances the model's explainability.

\begin{table}
    \centering
    \begin{tabular}{cc}
    \toprule
    
         Configuration&  Accuracy\\
         
    \midrule
    
        frozen backend, $\lambda = 0$ &  0.6960 \\

        frozen backend, $\lambda = 32$ &  0.8362 \\

        $\mu = 0$, $\lambda = 32$ &  0.8362 \\

        $\mu = 4$, $\lambda = 0$ &  0.8511 \\

    \midrule

        $\mu = 4$, $\lambda = 32$ &  \textbf{0.8573} \\
        
    \bottomrule
    \end{tabular}
    \vspace{4pt}
    \caption{Comparison of different key component configurations. Accuracy is reported.}
    \label{tab:Ablation}
\end{table}

\textbf{Effectiveness of Key Components}: To evaluate the efficacy of the MSE gaze constraint and leaky feature fusion, we conducted experiments using the Wild-DF dataset with several distinct configurations. The accuracy for different configurations is presented in Table \ref{tab:Ablation}. refers to the direct copying of gaze-related feature representations extracted by the pre-trained gaze backend ($f_{gb}(x^{i}_{d})$) to the corresponding elements within the representation extracted by the DeepFake detection backend, denoted as $\mathbf{r}|^{i}_{d_{1:\eta}}$. This table reveals that our proposed MSE gaze constraint and leaky feature fusion significantly enhance the model's accuracy. Notably, enabling leaky feature fusion with the frozen backend results in a significant 14.02\% increase in accuracy, which underscores the limitations and deficiencies of exclusively relying on features extracted by a frozen backend trained on different datasets. Additionally, the 1.49\% improvement observed when enabling the MSE gaze constraint further accentuates the model's enhancement through the integration of gaze-related features.

\section{Conclusion}

In this paper, we introduced the \textit{GazeForensics} framework, a novel approach that amalgamates gaze-related features with DeepFake detection, directing the model's attention to eye regions in the identification of face forgeries. Furthermore, we have presented an innovative biometric feature integration approach, incorporating MSE constraint and leaky feature fusion into the representation extracted by the CNN backend, resulting in a significant improvement in model accuracy and robustness. Our approach has achieved SOTA performance in the FaceForensic++ and WildDeepfake datasets, while concurrently exhibiting commendable accuracy in the Celeb-DF dataset. The regularization method we have proposed has broader applicability, as it can theoretically be adapted to most biometric-based DeepFake detection methods. We consider our work to be a significant advancement in the ongoing effort to combat DeepFake manipulation using biometric features.

\bibliographystyle{elsarticle-num}
\bibliography{reference}
\end{document}